# Fast Untethered Soft Robotic Crawler with Elastic Instability

Zechen Xiong[1], Yufeng Su[2], Hod Lipson[2]

*Abstract*—Enlightened by the fast-running gait of mammals like cheetahs and wolves, we design and fabricate a single-actuated untethered compliant robot that is capable of galloping at a speed of 313 mm/s or 1.56 body length per second (BL/s), faster than most reported soft crawlers in mm/s and BL/s. An in-plane prestressed hair clip mechanism (HCM) made up of semi-rigid materials, i.e. plastics are used as the supporting chassis, the compliant spine, and the force amplifier of the robot at the same time, enabling the robot to be simple, rapid, and strong. With experiments, we find that the HCM robotic locomotion speed is linearly related to actuation frequencies and substrate friction differences except for concrete surface, that tethering slows down the crawler, and that asymmetric actuation creates a new galloping gait. This paper demonstrates the potential of HCM-based soft robots.

## I. Introduction

Soft robots and manipulators offer many advantages compared to traditional robots, such as bio-compatibility, safety, resilience, and economic viability. These features make them an important method in a broad range of potential applications [1]–[3]. Drawing from the observation that most biological lifeforms have compliant bodies, soft robots use deformable and resilient material for both their torso and limbs. Popular materials like dielectric elastomer (DE), silicone rubber, and PDMS all have moduli of 10-1000 kPa and a density slightly over 1 g/cm$^3$, which are similar to those of human muscles. However, the high mass density and low energy density of these elastomers severely limit the speed and strength of soft robots.

To overcome the speed and force problem, researchers have proposed several solutions. Mosadegn et al. and Bas et al. [4], [5] suggest using improved geometry of the pneumatic network (pneu-net) to reduce the amount of gas needed for its inflation so that its response can be expedited. Such a proposal enables a fast actuation of tens of milliseconds yet at the cost of higher actuation pressure or energy source. Shepherd et al. [6], Bartlett et al. [7], Tolley et al. [8], and Keithly et al. [9] use explosive reactions in pneumatic networks to power jumping soft robots. This method generates fast and large leaps but is hard to iterate and control, and may cause damage to the matrix materials. Li et al. [10] and Wu et al. [11] use high-frequency electric power to drive fast-moving fish and terrestrial robots, but this process demands extra energy input, making it hard to remove the tether. Lin et al. [12] design a rolling caterpillar robot driven by a shape memory alloy (SMA) actuator, which moves at a speed exceeding 0.5 m/s.

Yet the robot lacks the ability to work continuously. Wang et al. [13] create a flower-shaped tiny magnetic soft robot that can wrap a living fly in 35 ms and deform at high frequencies but its working range is limited by the magnetic fields. Another solution is to use structural instability as a force amplifier to create rapid and strong soft robots [14]–[17], yet the reported techniques are restricted by the design and fabrication methods, impeding their practical use, especially in the untethered robotic field.

Inspired by the fact that most fast-moving animals have rigid body parts like spine and bones to support and build moving gaits, we investigate the potential of using semi-rigid materials, i.e., materials that are generally rigid enough to support weights yet can deform considerably at the same time, to create unique moving gait for untethered soft robots. Borrowing from the snap-through mechanism of a steel hair clip, we proposed a new strategy of using the in-plane prestressing of 2D materials like plastic shims and metal sheets to create bi-stable and multi-stable mechanisms[18]. Fig. 1A-1D illustrates the phenomenon that by pinning the two extremities of an angled plastic ribbon together, we obtain a hair-clip-like bi-stable mechanism that can be used for the propulsion of a fish robot (swimming at 436 mm/s or 2.03 BL/s, detailed research elaborated in [18]); if two such hair-clip mechanisms (HCM) are joined "head"-to-"head" (Fig. 1E-1F), a quadri-stable compliant structure can then be assembled from 2D materials with the strategy of in-plane pre-stressing. To make the system simpler and more controllable for the untethered biped robot, we have the two HCMs in its chassis as symmetric as possible and treat it as a bi-stable mechanism. In this way, the HCM biped robot can switch between two states—flexion (Fig. 1F) and extension—through mechanic actuation, which is achieved by a single servo in our case, and can perform a biomimetic galloping like a cheetah or a wolf does. On the other hand, if we make the two HCMs slightly different and actuate them separately, a completely different locomotion gait can be triggered. This is discussed in Section IV.

Experiments show such an untethered HCM biped soft robot with a single mini servo as actuation can locomote at a speed of 1.56 BL/s (313 mm/s), faster than most untethered soft robots. This work would be presented as follows. In section II, we explain the theory of HCMs, giving the theoretic solutions to the deformation and energy properties in the processes of assembly and buckling. In section III, we describe the materials, designs, and fabrication for the

*Research supported by the Dept. of Earth and Environmental Engineering, Columbia University.

[1]Zechen Xiong is with the Dept. of Earth and Environment Engineering at Columbia University, New York, NY 10027 USA (phone: 9173023864, e-mail: zx2252@columbia.edu).

[2]Yufeng Su and Hod Lipson are with the Dept. of Mechanical Engineering at Columbia University, New York, NY 10027 USA (e-mail: ys3399@columbia.edu and hl2891@columbia.edu).

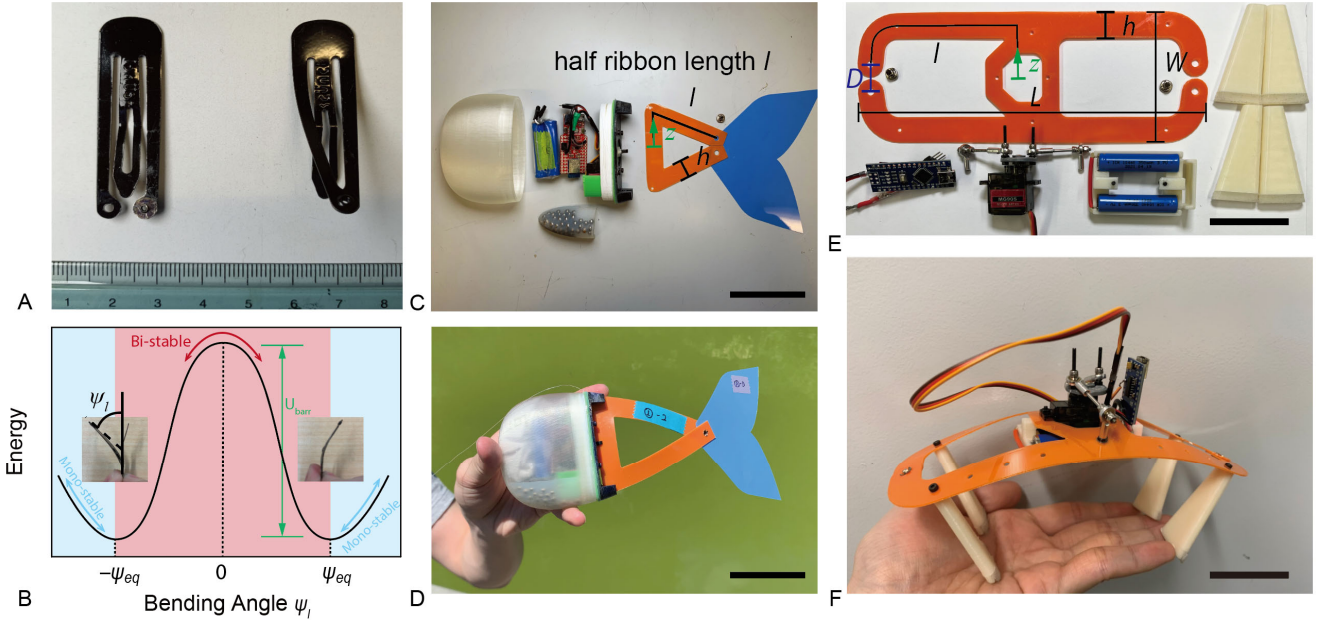

Figure 1. Principle of the hair-clip mechanism (HCM) and its applications to untethered soft robots. (A) and (B) The structure and energy landscape of a bi-stable steel hair clip, respectively. (C) and (D) The components and assembly of the untethered robotic fish with uni-sided triangular HCM as fish tail [18]. The fish swims at a speed of 2.03 BL/s (436 mm/s) (E) The components of the untethered biped compliant robot with a bilateral HCM chassis. Dimensions of the dual-HCM plastic chassis are $l$ = 129.1 mm, $L$ = 200.6 mm, $W$ = 75 mm, $D$ = 16 mm, $h$ =15 mm, and sheet thickness $t$ = 0.381 mm. (F) The assembled HCM bi-stable robot in its flexion state with a total length of $L_f$ = 176.5 mm. Scale bar = 5 cm.

untether biped robots. In section IV, we demonstrate the applications of HCMs in an untether and a tether soft robot and compare it to existing studies. In section V, we offer a conclusion.

## II. WORKING PRINCIPLES

The fast morphing and force amplifying effect of the HCM are from the snap-through buckling of bi- and multi-stable structures. It is observed that the lateral-torsional buckling of the ribbons when pinning the extremities together would lock the structure into a dome-like configuration with increased rigidity and bi-stability, which explains the functions of the HCM as a load-bearing skeleton and a high-speed actuator. Using the knowledge from Euler beam buckling theory and previous studies [18], [19], it is believed that the theoretic buckling shape can be given as

$$\varphi = \sqrt{l-z}\, A_1 J_{1/4}\left(\frac{1}{2}\sqrt{\frac{P_{cr}^{2}}{EI_\eta C}}(l-z)^2\right), \quad (1)$$

in which $\varphi$ is the lateral rotational angle, $l$ is the half ribbon length of an HCM (Fig. 1E), $z$ is the coordinate along the path of the ribbon (Fig. 1E), $A_1$ is a non-zero integration constant that can be determined from energy conservation, $J_{1/4}$ is the Bessel function of order ¼, indicating that the ribbons buckle in the shape of a Bessel function theoretically during the assemble, $P_{cr}$ is the critical load of the lateral-torsional buckling of the ribbon, $EI_\eta = Eh^3t/12$ is the out-of-plane bending stiffness of the ribbon, $C = Ght^3/3$ is the torsional rigidity of the rectangular-section ribbon, $E$ is Young's modulus, $G$ is the shear modulus of the sheet material, $h$ is the ribbon width and $t$ is the thickness of the sheet used in Fig.1E.

Plugging in the boundary condition of the assemble HCM, the value of the $P_{cr}$ can be derived as

$$P_{cr} = \frac{5.5618}{l^2}\cdot\sqrt{EI_\eta C}. \quad (2)$$

The three most significant factors that influence the kinematic and dynamic features of these HCM structures are the tip bending angle $\psi_l$, the energy barrier $U_{barr}$ (Fig. 1B), and the timescale $t_*$ of the snap-through process, which can be expressed as follows, respectively, according to the theory,

$$\psi_l = \frac{P_{cr}}{EI_\eta}\int_0^l \varphi(l-z)\,dz, \quad (3)$$

$$U_{barr} = 3P_{cr}\cdot D, \quad (4)$$

and

$$t_* = \frac{(2l)^2}{t\sqrt{E/\rho_s}}, \quad (5)$$

where $D$ is the prestressing distance (Fig. 1E) and $\rho_s$ is the density of the material.

The relation between the geometry of the unassembled HCM plate and the performance factors $\psi_l$ and $U_{barr}$ are demonstrated in Fig. 2A and 2B. The measured value of $\psi_l$ and $U_{barr}$ in the case of $l$ = 129.1 mm and $D$ = 16 mm are 0.44 rad and 43.4 mJ, and the theory gives corresponding values of 0.53 rad and 28.2 mJ, which are 20 % and −35 % in error compared with the experiments (Fig. 2). The considerable error is due to the angled shape of the ribbons that deviates from an encastred beam used to develop the theory and the path dependency and friction effect that make the measured energy barrier larger than the theory.

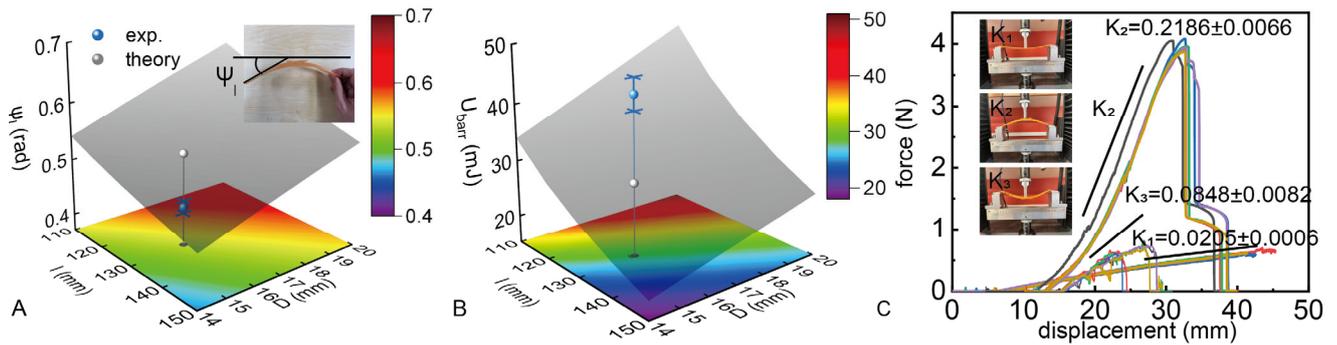

Figure 2. Theoretic solutions and stiffness properties of HCMs. (A) Theoretic tip bending angle $\psi_l$ w.r.t. half ribbon length $l$ and locking displacement $D$. (B) Theoretic energy barrier $U_{barr}$ w.r.t. $l$ and $D$. (C) Three-point bending tests showing the rigidities of the unassembled plastic dual-HCM chassis, the assembled plastic dual-HCM chassis in bi-stable direction, and the assembled plastic dual-HCM in mono-stable direction. The energy barrier $U_{barr}$ of the HCM chassis is calculated from the areas under the $K_2$ curves.

The design factors $l$ and $D$ from Eq. (3) and Eq. (4) are extracted to show how we can configure the energy and kinematic property to optimize the robotic performance (Fig. 2). It is observed that increasing the half ribbon length $l$ will decrease both the tip bending angle $\psi_l$ and the energy barrier $U_{barr}$, thus decreasing both the foot swinging amplitude in the galloping gait and the energy released in each swinging, leading to reduced mobility of the biped robot [14]; on the other hand, increasing the locking displacement $D$ will result in the opposite effects. Obviously, increasing the energy barrier of the HCM chassis of the robot would require a stronger yet larger and heavier servo motor for its actuation, making the designing process a trade-off.

Thin plastic plates are generally too compliant to work as chassis and support the weight of a robot. Yet with the technique of HCM, we manage to elevate the rigidity by in-plane prestressing. The effects are demonstrated with a three-point bending test in Fig. 2C. The unassemble plastic HCM chassis has a constant rigidity of $K_1 = 0.0205 \pm 0.0006$ N/mm, while the assembled HCM chassis has an equivalent rigidity of $K_2 = 0.2186 \pm 0.0066$ N/mm in the bi-stable direction and constant rigidity $K_3 = 0.0848 \pm 0.0082$ N/mm in the mono-stable direction (Fig. 2C), which are about 11 times and 4 times larger, respectively than the unassembled one. With a self-weight of 72 g of our self-contained robot, ignoring the mass distribution, the static deflections will be 34 mm, 3.2 mm, and 8.3 mm in these three scenarios, respectively, showing the rigidity advantage of the HCMs.

Another trait of the dual-HCM chassis is that the two HCMs in the structure can be tuned and actuated separately based on the theory illustrated above, thus creating more special gaits other than the gallop. Essentially, the dual-HCM chassis is a quadri-stable structure. And depending on dimensions like $l$'s and $D$'s of the two HCM, we can make it to be symmetric quadri-stable or asymmetric quadri-stable. The experiments based on this technique are carried out in Section IV.

### III. FABRICATION AND MEASUREMENTS

The dual-HCM untethered bi-stable biped soft robot is composed of the parts shown in Fig. 1E (for clarity, we refer to the untethered HCM robot as a bi-stable robot instead of a quadri-stable robot), including a laser-cut (Thunder Laser, Nova51) PETG plastic HCM spine (McMaster-Carr, 9536K) that also works as the chassis, an Arduino nano microcontroller board, an MG90s servo motor, a 3D-printed horn (Ultimaker S3, PLA) that connects the servo with the links to actuate the bi-stable HCM, a 3D-printed battery holder, a battery set with two 3.7 V and 350 mAh Li-ion cylindrical cells (AA Portable Power Corp., LC-10440, AAA size), two press-fit rivets (McMaster-Carr, 97362A), a few M2 screws and Nylock nuts, and four 3D-printed feet with silicone rubber (Smooth-On Inc., Ecoflex 10) glued to them (using Loctite Super Glue), which will be further discussed below.

We use semi-rigid material PETG plastic sheets to fabricate HCMs (Young's modulus $E = 1730$ MPa, measured in [18]) to develop the force-amplifying spine and supporting chassis of the bi-stable robot. Although not as soft as the classic soft materials like silicone rubber, HCMs preserve the most important features of soft robots—easy to design and fabricate, safe for humans, resilient to harsh conditions, and a continuum body that has infinite degrees of freedom.

Like the galloping of animals in nature, the forward marching of the robot requires anisotropic friction force. Since our single-servo and single-joint biped robot does not have sophisticated feet and soles, we employ a simple way of creating forward friction: the feet of the robot are made of half plastic (white color PLA) and half rubber (highlighted in grey), as shown in Fig. 3A. The friction coefficients on clean and smooth surfaces like wood and glass and coarse surfaces like marble floor and concrete surface are measured through dragging test and the difference of the two kinds of contact are illustrated in Fig. 3B. All averaged values and standard deviations are calculated from 5 measurements. The silicone rubber pads offer a 140%~400% increase in friction according to our measurements. To make the robot more stable in the lateral direction during locomotion, the feet are designed to have a thickening shape.

The measurement of robotic speed and displacement is carried out using videos by comparing the covered distance to the dimension of the robot itself. The tip bending angles are measured from pictures. The rigidity of the assembled HCM plate and the unassembled plastic plate are measured using a

3-point bending test with a span of 18 cm, which is shown in Fig. 2C (SUNS, UTM4000 tensile machine). The energy barrier of an assembled dual-HCM plate is calculated from the area under the load-displacement curve in Fig. 2C. Both the rigidity and the energy barrier value are averages of 6 measurements, including 3 times on each side of the plate, avoiding the errors brought by fabrication methods like laser cutting.

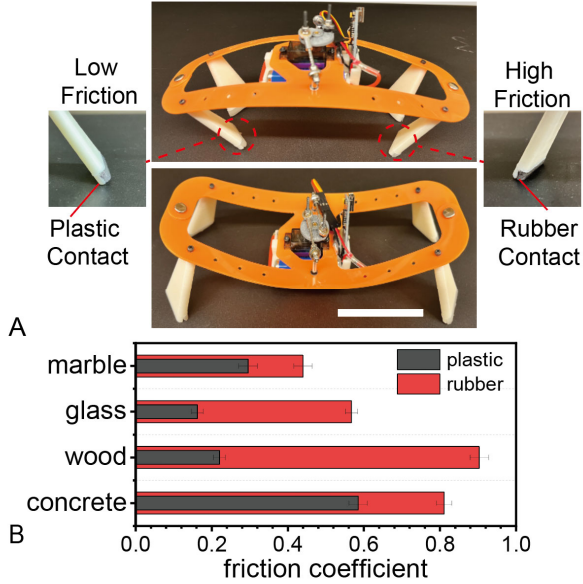

Figure 3. The feet and the contact conditions of the biped robot. (A) The directional locomotion enabled by friction asymmetry. The silicone rubber (Ecoflex) pads glued to the feet give asymmetric frictional forces (plastic contact gives low friction and rubber contact high friction) and convert the symmetric deformation of the system into directional locomotion. (B) Friction coefficient of the two contacts on different substrates. The tests show a 140~400% increase in the friction coefficient comparing silicone rubber contacts with plastic contacts, depending on the texture of the substrate. Scale bar = 5 cm.

## IV. APPLICATIONS

To demonstrate the functionality of the proposed strategy of using pre-stressed plastic HCMs as skeleton and force amplifiers in robotic propulsion, the kinematics of a self-contained dual-HCM biped robot and a tethered counterpart is studied in this section.

### A. Self-contained dual-HCM biped robot

The composition of the self-contained biped soft robot is illustrated thoroughly in Fig. 1E. It moves forward by snapping against the substrate and generating anisotropic friction through the silicone rubber pads on the feet. With a minimalistic design, the bi-stable HCM crawler shows a high locomotion speed with the experiment illustrated in Fig. 4A. On a flat and clean wood substrate, it can achieve a linear galloping speed of 1.56 BL/s, or 313 mm/s, during the video time length of about 2s. The video frames of the HCM biped robot show that it moves with a gait similar to that of some vertebrates like cheetahs and wolves when they are running. Affected by gravity of the robot, the upward bending (contraction in length) and downward bending (extension in length) is unequal. The upward bending of the dual-HCM spine (highlighted in white in Fig. 4B-4D) not only needs to overcome the energy barrier of the HCM chassis but also needs to vertically lift the robot, and thus have less kinetic energy built up. The process is usually featured by a very low horizontal speed and a comparatively higher vertical speed (Fig. 4B and 4C). The feet of the robot can be shortly off the ground (~70 ms, highlighted in grey in Fig. 4B-4D), just like the galloping gait of cheetahs. The downward bending of the dual-HCM spine (highlighted in blue in Fig. 4B-4D), however, borrows strength from the potential energy, thus has a higher horizontal speed and enables the crawler to cover most of the distance.

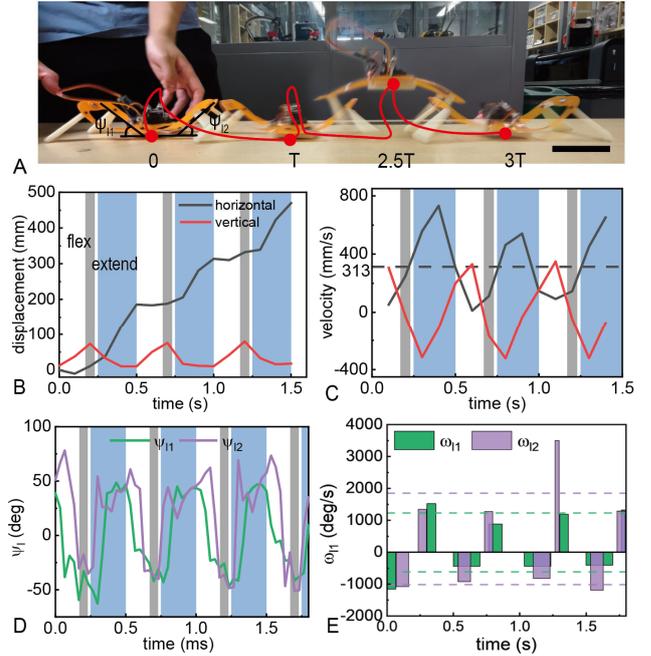

Figure 4. The kinematics of the directional galloping of the bi-stable biped robot under 2Hz actuation (period T = 500 ms). (A) Trajectory of the high-speed locomotion of the bi-stable HCM crawler in the video. An overall speed of 1.56 BL/s is obtained. (B) and (C) The displacement and velocity plots of the robot in horizontal and vertical direction in the video. Defining right direction as positive. The flexion and extension half-strides are distinguished by alternating white and blue. The hanging time is in grey. Most forward displacement takes place during the extension. (D) The angular displacement ($\psi_{l1}$ and $\psi_{l2}$) plots of the two ends of the dual-HCM spine in the video. Defining upward swinging as positive. (E) The angular velocity ($\omega_{l1}$ and $\omega_{l2}$) plots of the two ends of the dual-HCM spine during the snap-throughs. Defining upward swinging as positive. Averaged values of upward (extension) and downward (flexion) snapping of the two ends are given as dashed lines. A peak velocity of ~3500 deg/s is observed. Scale bar = 5 cm.

The snapping of the two HCMs in the chassis of the robot usually takes 70-200 ms according to our test, giving an angular velocity of 400-3500 deg/s (Fig. 4D and 4E) depending on the snapping direction. In comparison, a cheetah can stride up to 4 Hz during its fastest locomotion, where its angular velocity is approximated to be 1000 deg/s based on the literature and online video of cheetah running [20], [21]. These fast-swinging kinematics enable the HCM spine to exert high force against the ground and locomote the soft robot at a speed comparable to animals (1 to 100 BL/s) [14].

Several major factors influence the galloping speed of the robot. For example, the influence of actuation frequency of the single-servo driving system is almost linear, which means the higher the frequency the faster the robot runs, as shown in Fig. 5A. The influence of the substrates, however, may be more complicated. On smooth surfaces like wood, glass, and indoor marble floor, the speed has a positive correlation with the difference in friction coefficient between the two kinds of contacts (plastic and rubber contacts, Fig. 3); while on a coarse substrate like concrete, this anisotropic-friction-driving locomotion seems to fail (Fig. 5B) and the speed drop to 0 because of the roughness of the surface. With a more irregular and fluctuating texture of the surface, the plastic contacts between feet and substrate may generate more friction force than the rubber ones, thus neutralizing the anisotropic friction mechanism.

Note that despite the unconventional material used, the motion of the proposed crawler is supported and actively amplified by the continuous deformation of the prestressed HCM actuators under single-servo actuation. The stored energy in the elastically deformed HCMs, which can be tuned by changing the geometric dimensions, plays a dominant role in the force amplification, thus affecting the force output, velocity, and energy efficiency of the robot. It is reported that a more energy–stored bi-stable mechanism generates a larger energy gap, thus offering a larger force exertion and higher speed [14]. However, the HCMs of higher energy barriers increase the energy consumption of the system because they require a higher working voltage and a stronger servo motor to overcome the gap. Therefore, a trade-off should be considered when selecting the prestressing level of the HCM.

## B. Tethered dual-HCM biped robot

To further study the potential of the proposed HCM strategy in soft robotics, we design and fabricate a similar tethered biped robot as in Fig 6. After removing the battery and controlling system from the chassis, the robot achieves a total weight of 41 g, which is 57% of its self-contained counterpart. Experiments show that this robot can make use of two completely different gaits to move forward, depending on whether one or two HCMs are triggered during the actuation. This is because the fore-HCM ribbon is angled in the central area of the robot to support the servo, making it slightly stiffer and harder to buckle than the rear-HCM. By carefully configuring the rotation range of the servo horn, we can locomote the robot in either a sliding-and-jumping pattern with only one HCM buckling or a galloping mode with both the HCMs buckling. We term the motion sliding-and-jumping pattern because the robot will slide forward and take a small leap at the end of each period to recover, which is shown in Fig. 6B. The reason why the untethered robot generally does not have this trait is that its larger weight makes it hard to only buckle one side of the chassis.

The speed of these two moving patterns is measured to be 1.28 BL/s and 0.96 BL/s for the sliding-and-jumping mode and the galloping mode, respectively. Interestingly, although the robot is 43% underweight, the tethered robot is slower than the untethered robot. This is because the friction difference is also proportional to the weight and the light weight of the tethered robot leads to more vibration and collision that make the robot harder to maintain a stable motion thus lowering the speed. Therefore, the HCM technique is more suitable for untethered robots. It is also interesting to see that the tethered robotic gait with only one

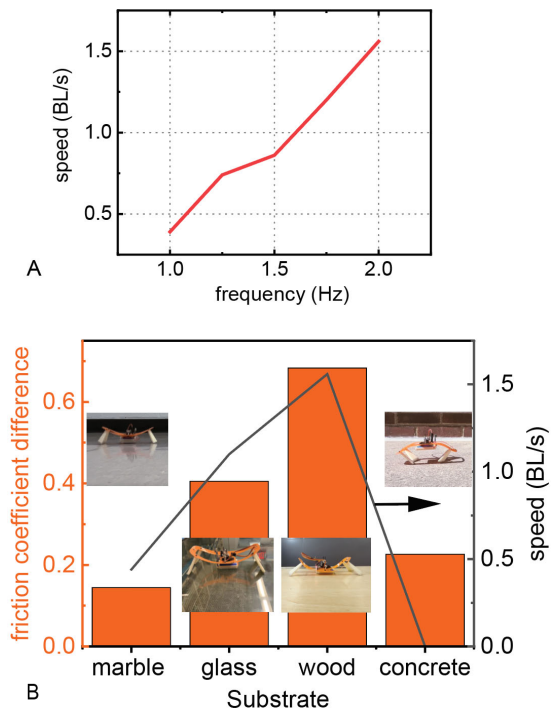

Figure 5. The factors that influence the speed of the HCM biped crawler. (A) The galloping speed of the robot is almost proportional to the actuation frequency. (B) The influence of substrates. On smooth surfaces like marble floor, glass, and wood, the locomotion speed of the robot is positively correlated with the difference of friction coefficient of the two kinds of contact (Fig. 3) on respective surfaces; while on coarse substrate like concrete, the anisotropic friction mechanism fails and the speed drop to 0.

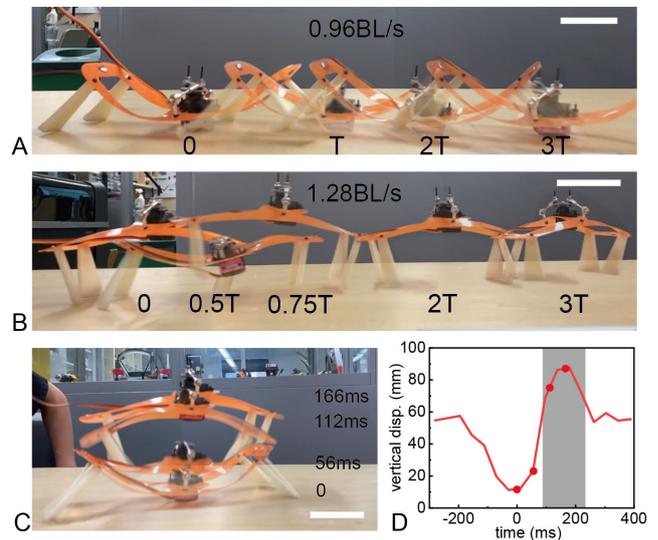

Figure 6. Galloping, sliding-and-jumping, and high jump sequence of tethered biped robot. (A) Trajectory of the locomotion of the tethered biped robot sharing the same galloping gait as the untethered robot. An overall speed of 0.96 BL/s is obtained. (B) Trajectory of the locomotion of the tethered biped robot where only the rear-HCM is actuated. An overall speed of 1.28 BL/s is obtained. (C) and (D) The high jump of the tethered robot. The frames in (C) correspond to the red dots in (D). Grey highlight represents the off-ground period for 146 ms. A off-ground height of 34 mm is observed. Scale bar = 5cm.

HCM buckling is faster than the normal galloping gait with both the HCMs buckling. The high jump gait indicates that the tethered robot can jump off the ground for 146 ms (58% of the total time) and 34 mm in each actuation during the galloping mode (Fig. 6C and 6D), while the sliding-and-jumping mode keeps the forefeet of the robot in contact with the ground most of the time, reducing vibration and collision and maintaining a more stable movement. With the multi-stable feature of the pre-stressed HCMs, more intriguing gaits can be designed.

*C. Comparisons*

Many previous studies have drawn people's attention to the fast and strong locomotion of soft robots. Fig. 7 shows the comparisons of the HCM-based crawler in this work with a few representatives of untethered on-land soft robots [22]–[27] in the categories of body length per second and mm per second, respectively. These soft-bodied crawlers, mostly having continuous compliant bodies and mono-stable structures, demonstrate a speed in the range of 0.45 to 40 mm/s or 0.008 to 0.6 BL/s because of either slow actuation speed or weak force exertion of the robots. However, Using semi-rigid 2D material as the compliant body as well as the force amplifier of the crawler, the proposed HCM soft robot is faster than most soft crawlers in mm/s and BL/s (Fig. 7). Another type of compliant robots using smart composite microstructure and motor-gear system also achieved high speed [28]–[31]. Meanwhile, the HCM-based compliant robots require only 2D fabrication and easy assembly due to the benefit of the in-plane prestressing.

## V. CONCLUSION

In this work, we propose a method to design rapid soft locomotive robots by prestressing 2D materials like plastic sheets, which we termed hair-clip mechanisms or HCMs due to their inspiration from the hair clips. We derive the design principles using the theoretic solutions obtained from previous work and fabricate a quadri-stable biped HCM crawler with two HCMs connected head-to-head. A running speed of 313 mm/s or 1.56 BL/s in untethered terrestrial soft robotics is measured in the experiments. The influence brought by multiple factors, including actuation frequency, substrates, tethering/untethering as well as symmetric/asymmetric actuation is discussed based on various experiments.

In the future, we are going to explore the compatibility of HCM with smart actuation methods like pneumatic actuation, shape memory alloy, and dielectric elastomer. The simple structure and fabrication of HCM robots may be useful. In the meanwhile, we will make HCM-based actuation systems with different materials and more capable designs.

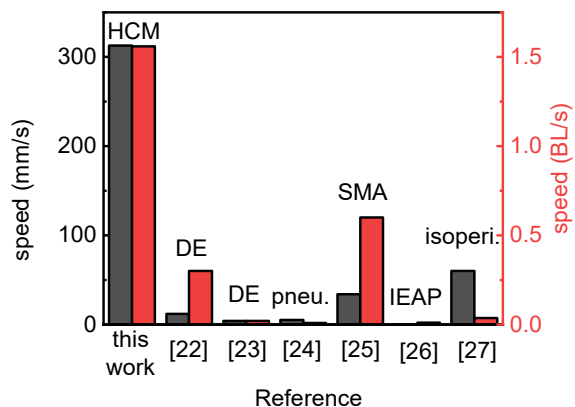

Figure 7. Comparison of locomotion speed between our proposed HCM biped robot and other reported untethered locomotive soft robots. Technologies used including dielectric elastomer (DE), pneu-nets, SMA, ion elecroactive polymer (IEAP), isoperimetric shape change.